\documentclass{article}

\usepackage{amssymb}
\usepackage[preprint]{neurips_2026}

\usepackage[utf8]{inputenc} 
\usepackage[T1]{fontenc}    
\usepackage{hyperref}       
\usepackage{url}            
\usepackage{booktabs}       
\usepackage{multirow}
\usepackage{makecell}
\usepackage{amsfonts}       
\usepackage{nicefrac}       
\usepackage{microtype}      
\usepackage[table]{xcolor}  
\usepackage{graphicx}
\usepackage{adjustbox}
\usepackage{wrapfig}
\definecolor{mycolor}{RGB}{64,128,220}
\definecolor{refgray}{RGB}{245,245,245}
\newcommand{\metricup}{\textcolor{red}{$\uparrow$}}
\newcommand{\metricdown}{\textcolor{red}{$\downarrow$}}

\usepackage{pifont}

\newcommand{\cmark}{\ding{51}}
\newcommand{\xmark}{\ding{55}}

\title{Unified Video-Action Joint Denoising for Dexterous Action and Data Generation}

\author{
Dingrui Wang$^{1,2,}$\thanks{
Equal contribution; 
$\dag$: project lead
}\quad YuAn Wang$^{2,*}$ \quad Jinkun Liu$^{2,3,*}$ \quad Yue Zhang$^2$\quad Mattia Piccinini$^1$ \\
\textbf{
Yu Sun$^{2,\dag}$ \quad Johannes Betz$^{1}$ }
\vspace{3pt}
\\
$^1$Technical University of Munich \quad 
$^2$ByteDance \quad 
$^3$Tsinghua University \quad 
}
\begin{document}

\maketitle

\begin{figure}[ht]
\vspace{-16pt}
\begin{center}
	\includegraphics[width=1\linewidth]{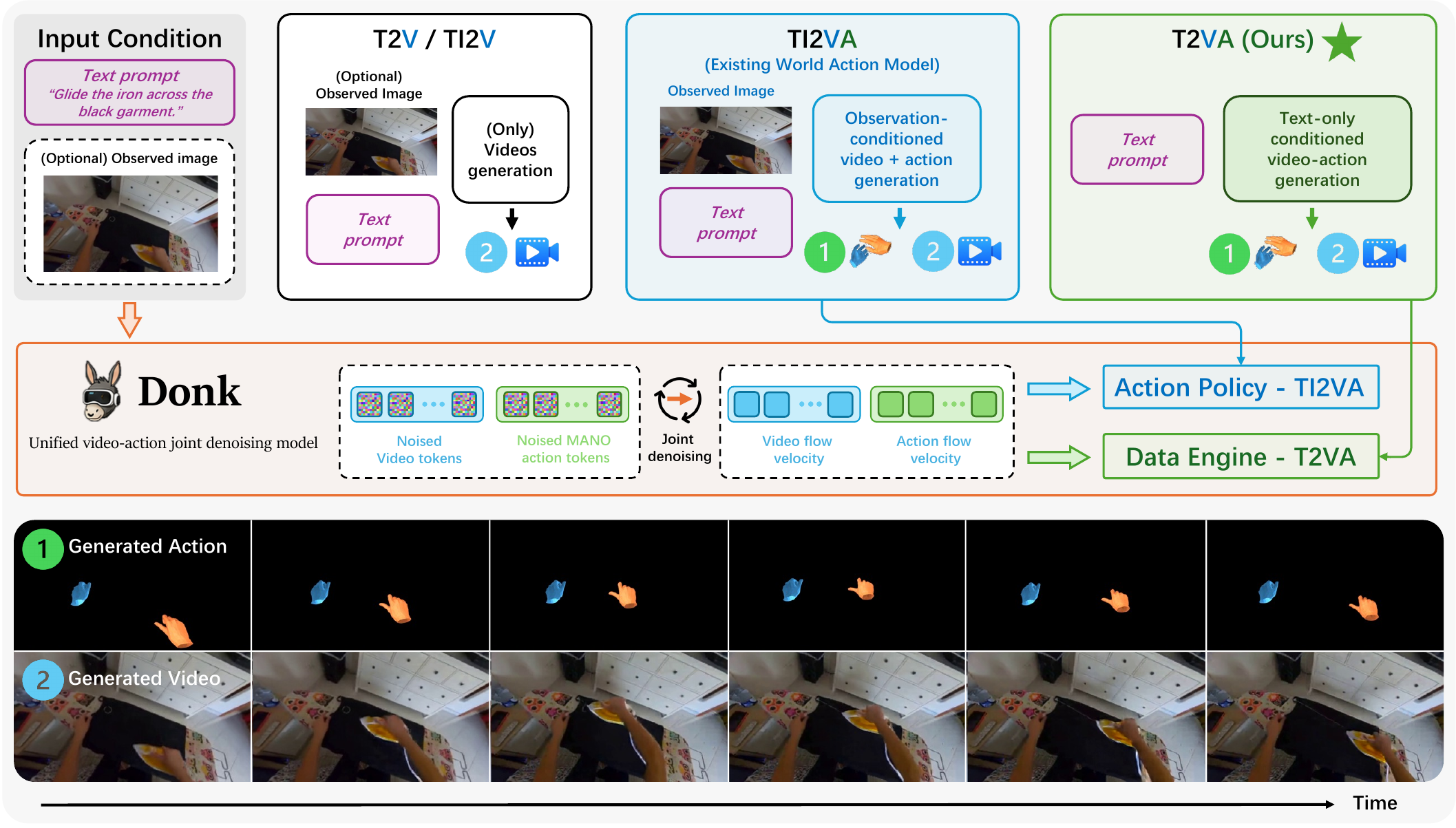}
\end{center}
\vspace{-7pt}
\caption{\small
\textbf{Donk unifies video-action generation.}
Given text alone, \textbf{Donk} generates paired interaction videos and spatio-temporally aligned MANO hand actions; with an observed image, \textbf{Donk} acts as an action policy.
}
\vspace{-2pt}
\label{fig:teaser}
\end{figure}

\begin{abstract}
Recent world action models leverage video foundation models by aligning broad
visual-dynamics priors with executable robot actions. We revisit this alignment
from a distributional perspective. Existing formulations typically narrow the
aligned prior into an observation-conditioned policy distribution over future
actions. In contrast, we keep the distribution broader by modeling the joint
space of interaction videos and executable hand trajectories under multiple
conditioning regimes. We propose \emph{Donk}, a unified video-action denoising
model for dexterous hands. With language, an initial image, and the initial hand
state, \emph{Donk} samples future videos and bimanual MANO trajectories as an
action policy. Without the image condition, the same denoising architecture
samples paired video-action rollouts from a text-conditioned distribution,
turning the aligned video prior into a data engine. Across action, video, and
text-only generation evaluations, \emph{Donk} improves dexterous trajectory
accuracy, preserves strong video fidelity, and produces smooth text-conditioned
action rollouts under the same unified training recipe.
\end{abstract}








\section{Introduction}

The success of vision-language models (VLMs) has motivated a growing line of
vision-language-action (VLA) policies that extend language and visual
understanding to robot control. The goal is to build general-purpose embodied
agents that can follow language instructions and perform diverse manipulation
tasks across objects, scenes, and embodiments. This is particularly challenging
for dexterous manipulation: to complete a language-specified task, an agent must
infer fine-grained hand-object interactions, reason about contact, anticipate
object motion, and produce temporally precise actions. Existing VLA policies have
made substantial progress by mapping language and visual observations directly to
robot actions
\cite{kim2024openvla,octoteam2024octo,black2024pi0,wen2025dexvla,zhong2025dexgraspvla,chi2023diffusionpolicy}.
However, action prediction alone treats actions primarily as output targets and
does not explicitly model the physical consequences of those actions. As a
result, the policy receives limited supervision about how the scene should evolve
under contact, even though such evolution is crucial for dexterous manipulation,
where small differences in hand pose, contact timing, and object motion can
determine task success.

World Action Models (WAMs) address this limitation by building action policies
on top of video foundation models (VFMs). Rather than predicting actions alone,
WAMs jointly predict future visual observations and actions, thereby coupling
motor commands with the visual futures they are expected to produce
\cite{zhu2025unifiedworldmodels,LiS-RSS-25,ye2026worldactionmodelszeroshot,bi2025motusunifiedlatentaction,yuan2026fastwam}.
This formulation is appealing for several reasons. First, VFMs pretrained on
large-scale heterogeneous video corpora provide rich spatiotemporal priors over
visual fidelity, temporal coherence, semantic controllability, human-object
interaction, contact dynamics, and object motion
\cite{wanteam2025wan}. Second, future video prediction provides a dense
supervisory signal beyond sparse action labels, encouraging the model to learn
physical regularities implicit in visual dynamics. Third, by aligning actions
with predicted visual futures, WAMs shift action learning from pure
state-action imitation toward video-action alignment, which can improve learning
efficiency and generalization, especially when robot data are limited or
heterogeneous.

Despite these advantages, existing WAMs are still primarily formulated as
observation-conditioned policies. Given a language instruction and the current
visual observation, they predict future observations and the corresponding action
trajectory. As illustrated in Fig.~\ref{fig:teaser}, in this sense, current WAMs can be viewed as
text-image-to-video-action (TI2VA) models: they condition on both text and an
initial image to generate aligned future videos and actions. However, this
observation-conditioned formulation captures only one instance of a broader
conditional video-action generation problem. From a probabilistic perspective, existing WAMs model a conditional distribution
over future videos and actions given language and an initial observation,
i.e., $p(\mathrm{video}, \mathrm{action} \mid \mathrm{text}, \mathrm{observation})$.
However, the initial observation is only one possible conditioning context, rather
than an inherent requirement of video-action generation. This suggests a broader
formulation in which the same action-aligned generative model can operate under
different conditioning contexts. A natural question therefore arises: \textbf{Can such a
model serve not only as an observation-conditioned policy, but also as a
language-conditioned generator of robot-relevant video-action experience?}

We answer this question by formulating a unified text/image-conditioned
video-action modeling problem. Given a language instruction, with or without an
initial visual observation, the model generates both an interaction video and a
spatially aligned action trajectory, namely
$p(\mathrm{video}, \mathrm{action} \mid \mathrm{text}, \mathrm{optional\ observation})$.
When the initial image is provided, this formulation specializes to TI2VA and serves
as an observation-conditioned policy. When the image condition is absent, it
becomes text-to-video-action (T2VA), where the model generates paired
visual-action supervision directly from language. This unified view turns TI2VA
policy learning and T2VA data generation into two conditioning modes of the same
end-to-end generative model, rather than two separate pipelines. Such a
formulation is especially useful for dexterous manipulation, where paired robot
visual-action trajectories are expensive to collect due to the difficulty of
teleoperation, calibration, and fine-grained action annotation. By contrast,
large-scale human-object interaction videos and text-conditioned video priors are
abundant. A text-only T2VA branch can therefore transform the broad interaction
priors of video foundation models into structured, action-aligned supervision for
robot learning.

Realizing this unified formulation is nontrivial. The model must preserve the
visual generative capability of the pretrained VFM while also learning to produce
action trajectories that are spatially synchronized and semantically consistent
with the generated video. Naively injecting action prediction into video
generation can interfere with video token representations and degrade visual
quality
\cite{ye2026worldactionmodelszeroshot,bi2025motusunifiedlatentaction}.
Meanwhile, post-hoc pipelines that first generate or reconstruct videos and then
extract actions introduce brittle intermediate representations, temporal
misalignment, and error accumulation. The key challenge is therefore to learn
video and action generation jointly in a single model, while maintaining visual
fidelity, action accuracy, and video-action temporal correspondence.

To address this challenge, we propose \emph{Donk}, a unified video-action joint
denoising model for dexterous manipulation. \emph{Donk} is built on a video
diffusion transformer \cite{wanteam2025wan,peebles2023scalable} and jointly
denoises video tokens and action tokens under a flow-matching paradigm
\cite{lipman2022flow}. Actions are represented as sequences of MANO hand
parameters, providing a structured representation of fine-grained dexterous hand
motion. Under image-conditioned inputs, \emph{Donk} functions as a TI2VA policy,
predicting both future visual observations and aligned hand actions from the
current scene. Under text-only inputs, the same model functions as a T2VA data
engine, generating paired interaction videos and synchronized hand-action
trajectories from language instructions. By unifying these two modes within a
single joint denoising framework, \emph{Donk} learns video-action consistency
from observed trajectories and reuses the resulting action-aligned generative
prior for text-conditioned data generation.

Our contributions are threefold:
\begin{itemize}
    \item We formulate text-to-video-action (T2VA) generation for dexterous
    manipulation, where the goal is to synthesize paired interaction videos and
    spatially aligned hand-action trajectories from language alone. To the best
    of our knowledge, this is the first exploration of T2VA as a text-only data
    engine for dexterous manipulation.

    \item We propose \emph{Donk}, a unified video-action joint denoising model
    built on a video diffusion transformer. By jointly denoising video tokens and
    MANO hand-action tokens within a flow-matching framework, \emph{Donk}
    supports both observation-conditioned TI2VA policy learning and
    text-conditioned T2VA data generation.

    \item We demonstrate that the unified formulation is effective for both
    policy learning and data generation. As a TI2VA policy, \emph{Donk} obtains
    the best hand RMSE and wrist-trajectory errors on OakInk benchmark and holds a good video fidelity with 0.2992 in LPIPS. 
    As a T2VA data engine, it maintains a good video
    quality while generating spatially aligned and temporally
    synchronized MANO hand actions.
\end{itemize}
\section{Related Work}

\paragraph{Action-centric embodied policies.}
Vision-language-action (VLA) models bring semantic knowledge from large
vision-language backbones into robot control, from web-scale action-token
policies to open generalist robot policies and recent flow-based action models
\cite{brohan2023rt2,octoteam2024octo,kim2024openvla,black2024pi0,physicalintelligence2025pi05,physicalintelligence2026pi07, chen2026egoman}.
Diffusion and diffusion-transformer policies further show that denoising
objectives are effective for multimodal continuous control, high-frequency
action chunks, and bimanual manipulation
\cite{chi2023diffusionpolicy,liu2024rdt1b,nvidia2025gr00tn1,pertsch2025fast}.
Dexterous and cross-embodiment systems extend this line with embodiment-aware
training, human-centric action spaces, post-training, memory, and online
specialization
\cite{wen2025dexvla,zhong2025dexgraspvla,beingbeyond2026beingh05,han2026dexhil,torne2026mem}.
These methods are strong action predictors, but they primarily optimize the
control interface. The future visual consequence of an action is usually not a
first-class output that is generated and checked together with the trajectory.
Our work instead treats video and bimanual hand motion as two synchronized views
of the same dexterous future.

\paragraph{Video world models and human-video priors.}
A complementary line uses video generation or predictive world models as the
interface for planning, policy learning, or data generation. Early video-based
robot planners synthesize future observations and recover actions through
inverse dynamics or tracking, while more recent video models serve as policies,
subgoal generators, or sources of physical supervision
\cite{ha2018worldmodels,du2023unipi,du2023vlp,zhou2024robodreamer,bruce2024genie,bharadhwaj2024gen2act,liang2024dreamitate,liang2025video}.
For dexterous manipulation, large-scale human video is especially important
because robot hand data is expensive. Recent work converts egocentric human
activity into language, hand motion, spatial grounding, and action-relevant
pretraining signals, and uses human videos to learn dexterous world dynamics
\cite{li2025vitra,beingbeyond2025beingh0,feng2025vipavla,hoque2025egodex,goswami2025dexwm,gao2026dreamdojo}.
Latent-action approaches further show that unlabeled or weakly labeled videos
can yield compact action-relevant representations
\cite{ye2025lapa,gao2025adaworld,luo2026jala}. These works establish video as a
rich source of physical priors, but many still separate visual imagination from
executable action recovery, or keep the learned world model in an implicit
latent form. We build on their insight while exposing both the rendered future
and the aligned hand trajectory.

\paragraph{Unified video-action world models.}
The closest recent work studies world-action or video-action models that learn
future observations and actions together. Representative systems jointly denoise
video and action, learn shared video-action latents, or combine video backbones
with action decoders, causal interleaving, and cascaded video/action modules
\cite{zhu2025unifiedworldmodels,LiS-RSS-25,cen2025worldvla,bi2025motusunifiedlatentaction,ye2026worldactionmodelszeroshot,lingbotva2026,pai2025mimicvideo,ma2026dit4dit}.
Efficiency-oriented variants show that the value of video prediction can come
from training-time world supervision even when explicit future rendering is
reduced at deployment
\cite{yuan2026fastwam,ye2026gigaworldpolicy,beingbeyond2026beingh07}. A related
representation-learning thread predicts future structure in latent space rather
than pixels, including JEPA-style video and VLA world models
\cite{assran2025vjepa2,sun2026vlajepa,zheng2025flare}. Our goal is
complementary: we adapt a pretrained video diffusion transformer into a
single-stream video-action denoiser for dexterous hands. Under text-only or
first-image-conditioned inputs, the same backbone generates an explicit visual
rollout and a normalized bimanual action trajectory, grounded by structured hand
state, camera geometry, and rendered state maps.

\section{Method}
\label{sec:method}

\begin{figure}[t]
\centering
\includegraphics[width=1.0\columnwidth]{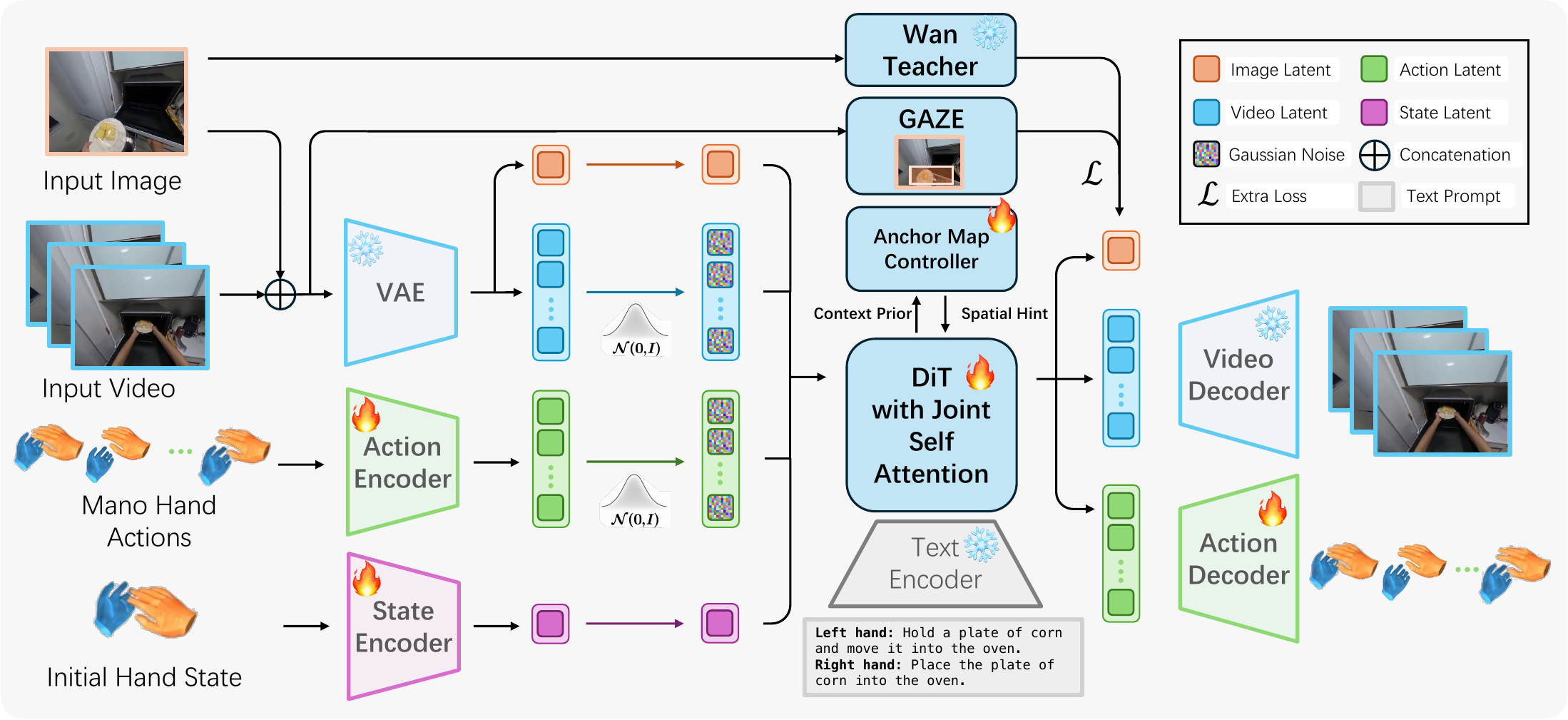}
\caption{Unified training framework.}
\label{fig:framework}
\end{figure}

\emph{Donk} is a unified video-action generative model for dexterous
manipulation. Given a language instruction and an optional initial image, it
generates an interaction video together with a spatially aligned MANO
hand-action trajectory. As shown in Fig.~\ref{fig:framework}, the same model
supports two conditioning modes: with the first image, it acts as a
text-image-to-video-action (TI2VA) policy; without the image, it acts as a
text-to-video-action (T2VA) data engine. Both modes share the same
hand-camera anchor interface, which provides a first-frame geometric scaffold
for projecting MANO hand states into the camera view and aligning generated
actions with the visual rollout.

\subsection{Unified Video-Action Modeling}

Let $c$ denote a language instruction, $V_{0:T}$ a manipulation video, and
$A_{1:T}$ the aligned future hand-action trajectory, where $T > 0$ is a fixed horizon. Donk models a unified
conditional video-action distribution:
\begin{equation}
  p_\theta(V_{0:T}, A_{1:T}\mid c, I_\star, g_0),
  \qquad
  I_\star\in\{I_0,\varnothing\},
\end{equation}
where $I_\star$ specifies the visual conditioning mode, and $g_0$ denotes the
initial hand-camera anchor. Specifically, $g_0$ contains the first-frame MANO
hand state and camera intrinsics, which determine how the hand geometry is
projected into the camera view.

In the TI2VA policy mode, the first image grounds the current scene and the
model predicts the future rollout:
\begin{equation}
  p_\theta^{\mathrm{policy}}(V_{1:T}, A_{1:T}\mid c, I_0, g_0)
  \triangleq
  p_\theta(V_{1:T}, A_{1:T}\mid c, I_\star=I_0, g_0).
\end{equation}
Here $g_0$ is obtained from the observed initial hand state and camera
intrinsics, providing the hand-camera configuration of the first frame.

In the text-only T2VA data-engine mode, no initial image is provided. The model
therefore generates the full interaction rollout from language and an initialized
hand-camera anchor:
\begin{equation}
  p_\theta^{\mathrm{engine}}(V_{0:T}, A_{1:T}\mid c,\tilde g_0)
  \triangleq
  p_\theta(V_{0:T}, A_{1:T}\mid c, I_\star=\varnothing, \tilde g_0).
\end{equation}
Here $\tilde g_0$ provides only a plausible first-frame geometric scaffold; it is
not a future action plan or trajectory-level condition. Under this formulation,
TI2VA policy learning and T2VA data generation are two conditioning modes of the
same video-action generator: the former uses an observed image and observed
initial hand-camera geometry, while the latter instantiates the missing initial
geometry before generation.

Videos are encoded into the pretrained Wan VAE latent space,
\begin{equation}
  x^\star=\mathcal{E}(V_{0:T}),
\end{equation}
and actions are represented as normalized continuous bimanual MANO trajectories,
$a^\star=A_{1:T}$, with invalid or missing hands masked during training.

\subsection{Joint Video-Action Architecture}

\textbf{Tokenization and Conditioning.}
Donk instantiates the unified distribution with a transformer denoiser
initialized from Wan2.2 TI2V-5B~\cite{wanteam2025wan}. The Wan stem patchifies
video latents into video tokens, and we add lightweight action and anchor
encoder to embed future MANO trajectories and the initial hand-camera anchor:
\begin{equation}
  z=[z^{\mathrm{video}},z^{\mathrm{action}},z^{\mathrm{anchor}}].
\end{equation}
The original Wan head predicts video outputs, while a lightweight action head
predicts MANO actions.

Image conditioning is injected in latent space. When $I_\star=I_0$, the
VAE-encoded first image replaces the first video latent frame and is assigned
timestep zero; when $I_\star=\varnothing$, this replacement is skipped. During
training, we drop $I_0$ with probability $0.30$, so both conditioning modes share
the same backbone, token layout, and objectives.

\begin{wrapfigure}[23]{r}{0.30\columnwidth}
    \vspace{-0.8em}
    \centering
    \includegraphics[width=\linewidth]{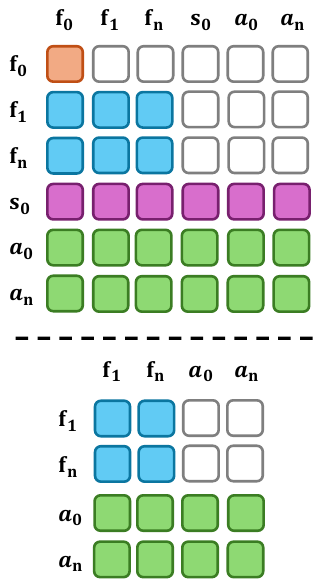}
    \vspace{-0.8em}
    \caption{Video-preserving attention mask.}
    \label{fig:attention}
    \vspace{0.6em}
\end{wrapfigure}

\textbf{Video-Preserving Joint Attention.}
A fully joint attention design would allow video tokens to attend to the newly
introduced action and anchor tokens, but this may disturb the pretrained video
generation prior. We therefore use a video-preserving attention mask: video
queries attend only to video tokens, whereas action and anchor queries attend to
the full sequence. As shown in Fig.~\ref{fig:attention}, this asymmetric
design keeps the visual stream close to the pretrained Wan computation, while
allowing action tokens to read both the generated visual rollout and the initial
hand-camera anchor. As a result, the model can align hand motions with the
evolving video without sacrificing the stability of the pretrained visual
generator.

\textbf{Anchor-Map Controller.}
Language and image conditioning alone provide limited direct control over the
image-plane location and pose of dexterous hands. We therefore introduce an
initial hand-camera anchor as an explicit geometric control signal for the first
frame. We denote this anchor as $g_0=(s_0,K)$, where $s_0$ is the initial MANO
hand state and $K$ denotes camera intrinsics. Given $g_0$, we render an anchor
map $M_0=\mathcal{R}(g_0)$ as a color-coded MANO skeleton image and encode it
into a latent anchor map $m_0=\mathcal{E}(M_0)$ using the frozen Wan VAE.

The latent anchor map is processed by a lightweight anchor-map adapter
$G_{\mathrm{anc}}$, which maps patchified anchor-map latents into the Wan token
space. Specifically, we patchify $m_0$ into anchor tokens and feed them into
$G_{\mathrm{anc}}$ to obtain a shared anchor-control representation $C$. For
each selected Wan layer $\ell\in\mathcal{S}$, a layer-specific MLP then produces
an anchor hint with the same token dimension as the first-frame video tokens:
\begin{equation}
  C = G_{\mathrm{anc}}(\mathrm{Patch}(m_0)),
  \qquad
  H_\ell = \mathrm{MLP}_\ell(C), \quad \ell\in\mathcal{S}.
\end{equation}
Here $C$ denotes a shared anchor-control representation, and $H_\ell$ is the
layer-wise anchor hint injected at layer $\ell$. We inject these hints through
gated first-frame anchor injection:
\begin{equation}
  z^{\mathrm{video}}_{\ell,0}
  \leftarrow
  z^{\mathrm{video}}_{\ell,0}
  +
  \gamma_\ell H_\ell,
  \qquad
  z^{\mathrm{video}}_{\ell,t>0}
  \leftarrow
  z^{\mathrm{video}}_{\ell,t>0}.
\end{equation}
The gates $\gamma_\ell$ are initialized to zero, so training starts from the
pretrained Wan behavior. Since $g_0$ specifies only an initial condition rather
than a future trajectory, the anchor hints are applied only to the first frame,
while hand-object evolution is learned by the joint video-action
denoiser.

In TI2VA, $g_0$ is obtained from the observed first-frame hand state and camera
intrinsics.In practice for prompt-only T2VA,
to ensure a reasonable initial hand pose and camera configuration, we train a
lightweight text-conditioned initializer to learn the empirical distribution of
first-frame hand-camera states and instantiate $\tilde g_0$. The initialized
state is used only to render the initial anchor map
$M_0=\mathcal{R}(\tilde g_0)$; the future interaction video and MANO trajectory
are still generated by the shared video-action denoiser.

\subsection{Training Objectives and Inference Modes}

Donk is trained with video-action flow matching, interaction-focused visual
supervision, and teacher-prior regularization. The denoiser predicts video and
action velocities $(\hat v_x,\hat v_a)$, supervised by the corresponding
flow-matching targets $(v_x,v_a)$.

The primary objective consists of video-flow matching and masked action-flow
matching:
\begin{equation}
 \mathcal{L}_{\mathrm{video}}
 =
 \|\hat v_x-v_x\|_2^2,
 \qquad
 \mathcal{L}_{\mathrm{action}}
 =
 \frac{
 \|M_a\odot(\hat v_a-v_a)\|_2^2
 }{
 \max(\sum M_a,1)
 },
\end{equation}
where $M_a$ masks invalid hand dimensions. To emphasize hand-object interaction
regions, we additionally use a hand-focused video loss
$\mathcal{L}_{\mathrm{gaze}}$, which weights video-flow errors around rendered
hand regions.

We also use a frozen Wan teacher prior to stabilize the visual generation path.
The teacher receives the same video latent and text condition
and predicts a video velocity $\hat v_x^{\,\mathrm{tea}}$:
\begin{equation}
  \mathcal{L}_{\mathrm{prior}}
  =
  \|\hat v_x-\hat v_x^{\,\mathrm{tea}}\|_2^2 .
\end{equation}
This term is applied only when the image condition is kept, preventing the
text-only branch from imitating an image-conditioned teacher without access to
the image.

The full denoiser objective is
\begin{equation}
 \mathcal{L}_{\mathrm{Donk}}
 =
 \lambda_v
 \left(
 \mathcal{L}_{\mathrm{video}}
 +
 \lambda_g\mathcal{L}_{\mathrm{gaze}}
 \right)
 +
 \lambda_a\mathcal{L}_{\mathrm{action}}
 +
 \lambda_p\mathcal{L}_{\mathrm{prior}} .
\end{equation}
The action and anchor interfaces, including the anchor-map adapter, are trained
end-to-end through this objective. For prompt-only T2VA, we separately train a
lightweight text-conditioned initializer to instantiate a plausible initial
hand-camera anchor; this auxiliary model is used only to provide the first-frame
geometric scaffold.

At inference time, Donk supports two modes. In TI2VA policy mode, it receives
$(c,I_0,g_0)$, clamps the first video latent using $I_0$, and generates the
future video-action trajectory. In T2VA data-generation mode, it receives only
the language instruction $c$. We instantiate a plausible initial hand-camera
anchor $\tilde g_0$, render its anchor map, and use the same shared denoiser to
generate the interaction video and synchronized MANO action trajectory. To
preserve the pretrained video prior, we freeze the text encoder, VAE, teacher
model, and most Wan blocks, and train only the action and anchor interfaces,
anchor-map adapter, action head, and a small subset of Wan layers.
\section{Experiments}

We train \emph{Donk} on VITRA-1M dataset~\cite{li2025vitra} with 64 NVIDIA Hopper GPUs with VRAM 96GB under PyTorch FSDP2~\cite{zhao2023pytorch}.
Each GPU processes one clip and the
effective batch size is 64 clips. We use bfloat16 precision and AdamW with a
constant learning rate $2\times10^{-5}$, default
$(\beta_1,\beta_2)=(0.9,0.999)$, $\epsilon=10^{-8}$, weight decay 0.01, and
gradient clipping at 1.0. 


\subsection{Action Accuracy for TI2VA}

\paragraph{Offline action accuracy.}
We first evaluate TI2VA on the OakInk2~\cite{zhan2024oakink2} first-person view benchmark. All methods
sample 10 futures per example. Following EgoMAN~\cite{chen2026egoman},
we use standard hand-trajectory metrics: Average Displacement Error (ADE), Final
Displacement Error (FDE), and Dynamic Time Warping (DTW), all reported in
meters, and wrist rotation error (ROT), reported in degrees. DTW-S and DTW-L are the short-window and open-end variants used
by the evaluator. ROT is the geodesic distance between predicted and
ground-truth wrist orientations. We additionally report hand RMSE for MANO
finger-pose accuracy. Lower is better for all metrics. For stochastic
prediction, we report best-of-$K$ results with $K\in\{5,10\}$.

\begin{table*}[t]
\centering
\caption{\textbf{Action policy model comparison on the OakInk2 first-person view benchmark.}
All metrics are lower-is-better. $K_5$ and $K_{10}$ denote best-of-$K$
evaluation.}
\label{tab:ti2va_action}
\vspace{2mm}
\footnotesize
\setlength{\tabcolsep}{3pt}
\renewcommand{\arraystretch}{1.08}
\begin{adjustbox}{max width=\textwidth}
\begin{tabular}{l|c|cc|cc|cc|cc|cc}
\toprule
\multirow{2}{*}{\textbf{Method}} & \multicolumn{1}{c|}{\textbf{Hand}} & \multicolumn{2}{c|}{\textbf{ADE}} & \multicolumn{2}{c|}{\textbf{FDE}} & \multicolumn{2}{c|}{\textbf{DTW-S}} & \multicolumn{2}{c|}{\textbf{DTW-L}} & \multicolumn{2}{c}{\textbf{ROT}} \\
\cmidrule(lr){2-2}\cmidrule(lr){3-4}\cmidrule(lr){5-6}\cmidrule(lr){7-8}\cmidrule(lr){9-10}\cmidrule(lr){11-12}
& RMSE\metricdown & $K_5$\metricdown & $K_{10}$\metricdown & $K_5$\metricdown & $K_{10}$\metricdown & $K_5$\metricdown & $K_{10}$\metricdown & $K_5$\metricdown & $K_{10}$\metricdown & $K_5$\metricdown & $K_{10}$\metricdown \\
\midrule
VITRA~\cite{li2025vitra} & 0.444 & 0.067 & 0.065 & 0.108 & 0.105 & 0.062 & 0.060 & 0.039 & 0.038 & \textbf{15.15} & \textbf{14.64} \\
Being-H0-1B~\cite{beingbeyond2025beingh0} & 0.587 & 0.118 & 0.107 & 0.131 & 0.120 & 0.118 & 0.107 & 0.098 & 0.090 & 40.52 & 38.18 \\
Being-H0-8B~\cite{beingbeyond2025beingh0} & 0.615 & 0.082 & 0.075 & 0.098 & 0.092 & 0.081 & 0.074 & 0.064 & 0.057 & 31.16 & 29.98 \\
DreamZero-alike & 0.262 & 0.062 & 0.057 & 0.100 & 0.094 & 0.059 & 0.054 & 0.040 & 0.037 & 20.48 & 19.00 \\
\rowcolor{mycolor!18}\textbf{Donk-TI2VA} & \textbf{0.238} & \textbf{0.055} & \textbf{0.049} & \textbf{0.090} & \textbf{0.079} & \textbf{0.052} & \textbf{0.046} & \textbf{0.032} & \textbf{0.029} & 16.05 & 14.95 \\
\bottomrule
\end{tabular}
\end{adjustbox}
\end{table*}
\vspace{-2mm}

Table~\ref{tab:ti2va_action} shows that \emph{Donk}-TI2VA gives the best hand
pose and wrist-translation results among the compared methods. The gains are
consistent across ADE, FDE, and both DTW variants, indicating better spatial
tracking over the full trajectory rather than only better endpoints. VITRA~\cite{li2025vitra} is
slightly better on wrist rotation, but \emph{Donk} remains close while being
substantially stronger on position and finger pose. Compared with the
Being-H0~\cite{beingbeyond2025beingh0} baselines, \emph{Donk} improves every reported metric.

\begin{table*}[t]
\centering
\caption{\textbf{TI2VA ablation with action-model metrics.}
All metrics are lower-is-better and use the same best-of-5 and best-of-10
selectors as Table~\ref{tab:ti2va_action}.}
\vspace{2mm}
\label{tab:ti2va_ablation}
\footnotesize
\setlength{\tabcolsep}{3pt}
\renewcommand{\arraystretch}{1.08}
\begin{adjustbox}{max width=\textwidth}
\begin{tabular}{l|cc|c|cc|cc|cc|cc|cc}
\toprule
\multirow{2}{*}{\textbf{Variant}} & \multicolumn{2}{c|}{\textbf{Cond.}} &
\multicolumn{1}{c|}{\textbf{Hand}} & \multicolumn{2}{c|}{\textbf{ADE}} &
\multicolumn{2}{c|}{\textbf{FDE}} & \multicolumn{2}{c|}{\textbf{DTW-S}} &
\multicolumn{2}{c|}{\textbf{DTW-L}} & \multicolumn{2}{c}{\textbf{ROT}} \\
\cmidrule(lr){2-3}\cmidrule(lr){4-4}\cmidrule(lr){5-6}
\cmidrule(lr){7-8}\cmidrule(lr){9-10}\cmidrule(lr){11-12}\cmidrule(lr){13-14}
& Gaze & State & RMSE\metricdown &
$K_5$\metricdown & $K_{10}$\metricdown &
$K_5$\metricdown & $K_{10}$\metricdown &
$K_5$\metricdown & $K_{10}$\metricdown &
$K_5$\metricdown & $K_{10}$\metricdown &
$K_5$\metricdown & $K_{10}$\metricdown \\
\midrule
Donk-TI2VA (full) & \cmark & \cmark & \textbf{0.238} & \textbf{0.055} & \textbf{0.049} & \textbf{0.090} & \textbf{0.079} & \textbf{0.052} & \textbf{0.046} & \textbf{0.032} & \textbf{0.029} & \textbf{16.05} & \textbf{14.95} \\
Donk-TI2VA (wo Gaze) & \xmark & \cmark & 0.258 & 0.058 & 0.053 & 0.093 & 0.082 & 0.055 & 0.049 & 0.035 & 0.032 & 18.17 & 16.21 \\
Donk-TI2VA (base) & \xmark & \xmark &  0.262 & 0.062 & 0.057 & 0.100 & 0.094 & 0.059 & 0.054 & 0.040 & 0.037 & 20.48 & 19.00 \\
\bottomrule
\end{tabular}
\end{adjustbox}
\end{table*}

\paragraph{Conditioning ablation.}
Table~\ref{tab:ti2va_ablation} isolates the conditioning used by the final
TI2VA model. ``Gaze'' denotes the Gaze module, and ``State'' denotes
the state expert. State conditioning alone already
improves the base model on all metrics. Adding the hand-focused cue gives the
full model, which further improves hand RMSE, trajectory error, and rotation
error. The trend is consistent across all metrics: the state expert is useful, and the gaze module gives an additional but
smaller gain.

\begin{figure}[t]
\centering
\includegraphics[width=1\columnwidth]{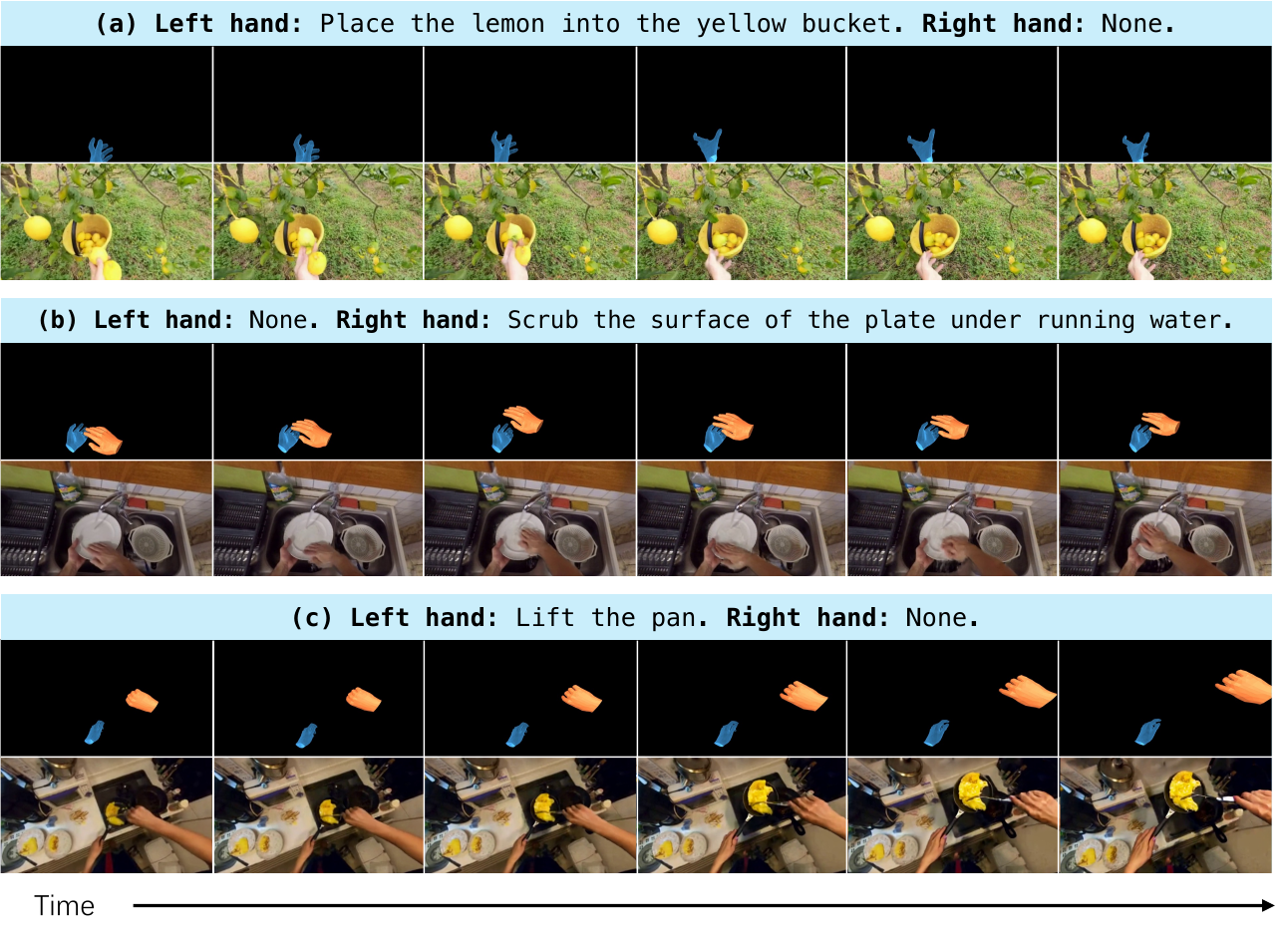}
\vspace{-2mm}
\caption{TI2VA alignment examples. Example (a) features part of the hand is missing at the beginning, while example (b) features hand occlusion and fluid interaction. Example (c) features height change and tool interaction.
}
\label{fig:ti2va_alignment}
\vspace{-1mm}
\end{figure}

\begin{table}[t]
\caption{\textbf{EgoDex aggregate metrics.}
All rows use the same protocol. Lower is better for LPIPS, tLPIPS, and FVD;
higher is better for PSNR, SSIM, CLIP-I, and CLIP-S.}
\vspace{2mm}
\centering
\footnotesize
\setlength{\tabcolsep}{3pt}
\renewcommand{\arraystretch}{1.08}
\begin{adjustbox}{max width=\textwidth}
\begin{tabular}{l|ccc|cc|c|c}
\toprule
\multirow{2}{*}{\textbf{Method}} & \multicolumn{3}{c|}{\textbf{Frame Fidelity}} & \multicolumn{2}{c|}{\textbf{Semantics}} & \multicolumn{1}{c|}{\textbf{Temp.}} & \multicolumn{1}{c}{\textbf{Dist.}} \\
\cmidrule(lr){2-4}\cmidrule(lr){5-6}\cmidrule(lr){7-7}\cmidrule(lr){8-8}
& PSNR\metricup & SSIM\metricup & LPIPS\metricdown & CLIP-I\metricup & CLIP-S\metricup & tLPIPS\metricdown & FVD\metricdown \\ 
\midrule
Wan2.2-TI2V-5B & \underline{19.50} & \underline{0.7855} & \underline{0.3061} & \underline{0.9119} & 0.2004 & 0.0429 & 81.87\\
Wan2.1-I2V-14B & 19.47 & 0.7742 & 0.3252 & 0.9090 & \underline{0.2053} & 0.0472 & \textbf{68.97} \\
Wan2.1-VACE-14B & 17.16 & 0.7220 & 0.4067 & 0.8551 & \textbf{0.2187} & \textbf{0.0197} & 103.85 \\
\rowcolor{mycolor!18}\textbf{Donk-TI2VA} & \textbf{19.84} & \textbf{0.7908} & \textbf{0.2992} & \textbf{0.9172} & 0.1982 & \underline{0.0340} & \underline{75.13} \\
\bottomrule
\end{tabular}
\end{adjustbox}
\label{tab:egodex}
\end{table}

\begin{table}[t]
\centering
\caption{\textbf{T2VA visual and semantic quality benchmark.}
Open video baselines are compared only on video metrics because they do not
generate paired action trajectories.}
\label{tab:t2va_visual_semantic}
\footnotesize
\setlength{\tabcolsep}{3pt}
\renewcommand{\arraystretch}{1.08}
\begin{adjustbox}{max width=\linewidth}
\begin{tabular}{l|c|cccc}
\toprule
\multirow{2}{*}{\textbf{Method}} & \multirow{2}{*}{\textbf{Input}} & \multicolumn{4}{c}{\textbf{Visual \& Semantic Quality}} \\
\cmidrule(lr){3-6}
& & FVD\metricdown & VLM judge\metricup & CLIP-S\metricup & tLPIPS\metricdown \\
\midrule
Wan2.2-5B-I2V & text & 306.2 & 1.59 & 0.2508 & \textbf{0.0147} \\
\rowcolor{mycolor!18}\textbf{Donk-T2VA} & text & \textbf{191.1}  & \textbf{2.37} & \textbf{0.2572} & 0.0215 \\
\bottomrule
\end{tabular}
\end{adjustbox}
\end{table}

\subsection{Video Quality for TI2VA}

\paragraph{EgoDex protocol.}
The action model must also preserve the world-modeling side of the task:
the generated video and generated action need to stay consistent. We use a LOME evaluation \cite{gao2026lome} based on EgoDex \cite{hoque2025egodex}
with 1000 samples, 17 frames, and 832$\times$480 resolution. PSNR, SSIM, and
LPIPS measure frame fidelity; CLIP-I and CLIP-S measure visual identity and
text-video alignment; tLPIPS measures temporal flicker; FVD measures generated
video distribution. 

Table~\ref{tab:egodex} shows that the action stream does not degrade the video
side. \emph{Donk}-TI2VA has the best PSNR, SSIM, LPIPS, and CLIP-I among the
matched runs. Pure
video baselines still lead on some video-only metrics: Wan2.1-I2V has the lowest
FVD, and Wan2.1-VACE has the best CLIP-S and tLPIPS. The main gain for
\emph{Donk} is that the video remains strong while the hand trajectory follows
the rollout. In Fig.~\ref{fig:ti2va_alignment}, the predicted actions track hand
motion and remain stable under partial hand occlusion and interaction.

\subsection{Visual and Semantic Quality for T2VA}

\begin{figure}[t]
\centering
\includegraphics[width=1\columnwidth]{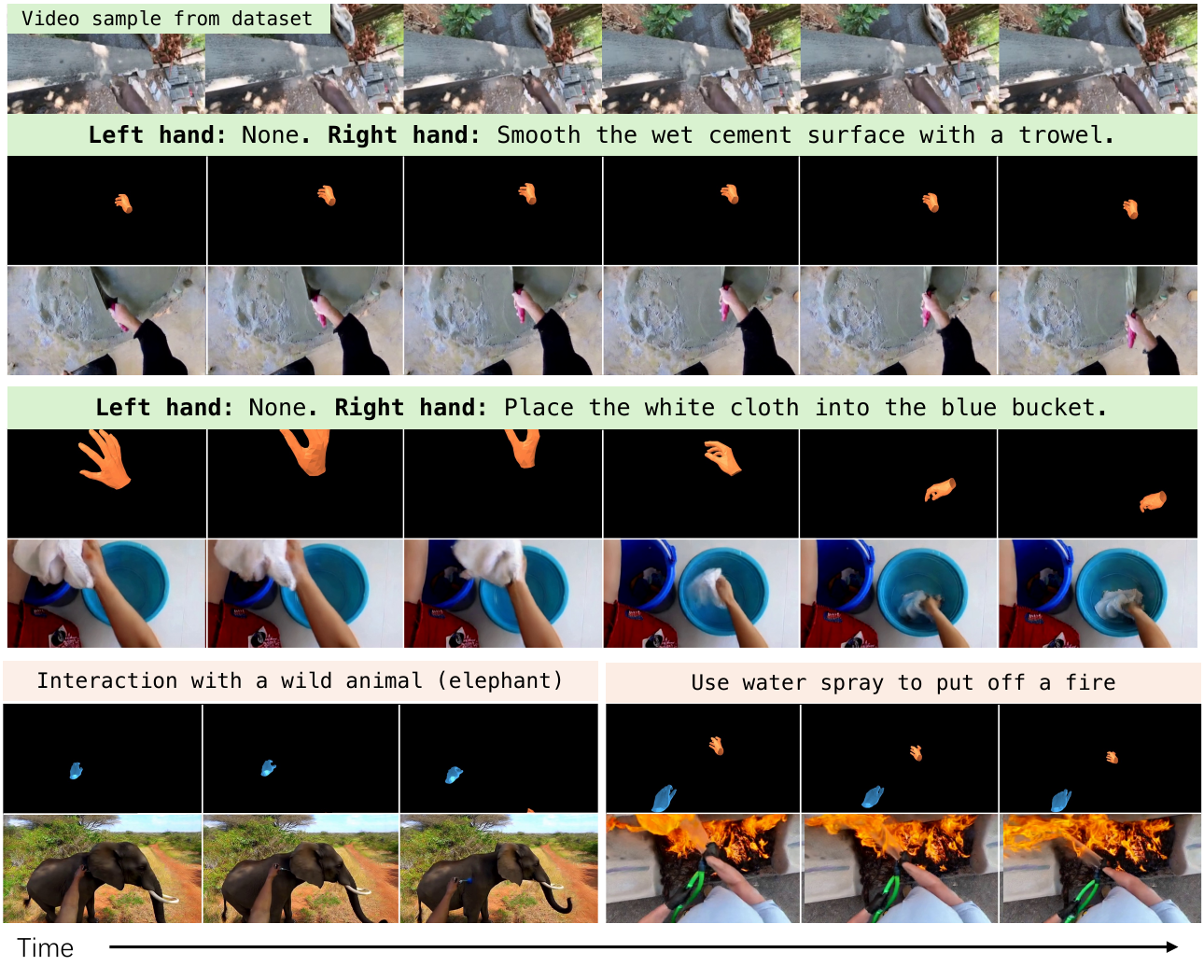}
\vspace{-2mm}
\caption{T2VA rollouts with only text as input.}
\label{fig:t2va}
\end{figure}

T2VA evaluates the second role of the same denoiser: generating paired
video-action data from text alone. We separate the video-quality comparison from
the action diagnostics because open video generation baselines provide strong
visual references but do not output executable bimanual actions.
Fig.~\ref{fig:t2va} shows that the text-only interface can sample paired
rollouts for prompts outside typical lab-collected manipulation distributions,
including outdoor animal interaction, emergency fire scenario and etc.
Table~\ref{tab:t2va_visual_semantic} shows that \emph{Donk}-T2VA maintains competitive visual and semantic quality compared to the off-the-shelf model.
It lowers FVD while improving the VLM judge score, where each generated video and its text instruction (100 samples from EgoDex are used) are sent to a VLM to evaluate instruction-following alignment on a 0-5 scale. The baseline does not output actions, so this
comparison only evaluates the video side.

\section{Conclusion}

We presented \emph{Donk}, a unified video-action joint denoising model for
dexterous world modeling. The central idea is to use the video-action alignment
learned by a World Action Model not only for observation-conditioned action
prediction, but also as the generative space for text-conditioned data creation.
With one Wan-initialized denoising backbone, \emph{Donk} supports TI2VA as a
policy-style action model and T2VA as a text-only video-action data engine,
sharing the same video latent space, bimanual action representation, geometric
state-map control, and flow-matching objective.

This unification changes the role of a dexterous WAM. Instead of training a
video model, an action model, and a data generator as separate systems,
\emph{Donk} makes action prediction and data synthesis two uses of the same
aligned prior. In TI2VA, the model achieves strong dexterous prediction results,
with clear gains in hand-pose accuracy and best-of-10 translational trajectory
metrics, while preserving competitive video quality and improving hand-action
following on the matched EgoDex-style evaluation. In T2VA, the same denoising
core generates paired video-action rollouts from text alone providing an initial path toward using WAMs directly as data engines. 


\newpage

\bibliographystyle{unsrt}
\bibliography{refs}


\end{document}